%% file: main.tex
\documentclass[11pt,a4paper]{article}

\usepackage[english]{babel}
\usepackage[utf8]{inputenc}

\usepackage{microtype}           
\usepackage[onehalfspacing]{setspace} 
\usepackage{parskip}             
\usepackage[hang,flushmargin]{footmisc} 

\usepackage[a4paper,margin=1in]{geometry} 
\usepackage{fancyhdr}
\usepackage{graphicx}
\usepackage{booktabs}
\usepackage{makecell}
\usepackage[section]{placeins} 
\usepackage{amsfonts}
\usepackage[singlelinecheck=false]{caption} 
\usepackage{threeparttable}
\usepackage[table]{xcolor}

\usepackage{amsmath,amssymb}

\usepackage{csquotes}
\usepackage{natbib}

\usepackage[hidelinks]{hyperref}

\usepackage[colorinlistoftodos]{todonotes}

\usepackage{titling}

\pretitle{\vspace{2em}\centering\Huge\bfseries}
\posttitle{\par\vspace{1em}}
\preauthor{\centering\large}
\postauthor{\par\vspace{0.5em}}
\predate{\centering\small}
\postdate{\par\vspace{2em}}

\title{Title of Your Report}
\author{Simon Polichinel von der Maase}
\date{June 7, 2025}

\begin{document}

\begin{titlepage}
    \centering
    \vspace*{3cm}
    
    {\Huge \textbf{The Currents of Conflict:\\} Decomposing Conflict Trends with Gaussian Processes\par}
    \vspace{1.5cm}
    {\Huge Working Paper}
    \vspace{1.5cm}

    \noindent\makebox[\textwidth]{\Large\textemdash\quad$\diamond$\quad\textemdash}
    \vspace{1.5cm}

    {\Large Simon Polichinel von der Maase \par}
    \vspace{0.5cm}
    \textit{PRIO -- Oslo} \\
    \textit{\today}
    \vspace{1.5cm}
    
\textbf{Author’s note}: This paper is based on material originally included in the author’s doctoral dissertation \citep{vondermaase2023}. It is currently being revised and extended as a standalone article, drawing on feedback and discussions with colleagues and reviewers, some of whom may be added as co-authors in future versions.

    \vfill
    \hspace{1cm}
    \includegraphics[width=0.25\textwidth]{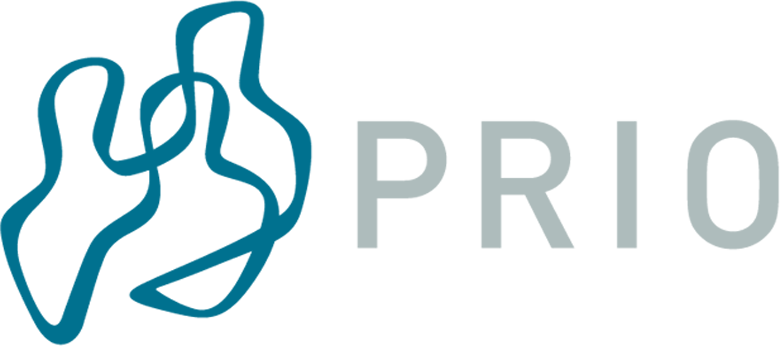}
    \hspace{1cm}

    \thispagestyle{empty}
\end{titlepage}


\input{sections/abstract}


\section{Introduction}
\input{sections/introduction}

\section{Theories, insights and derived modelling criteria}
\input{sections/theories}

\section{Data}

\input{sections/data}

\section{Gaussian processes}
\input{sections/gps}

\section{Combining the patterns}
\input{sections/combining}

\section{Evaluating the approach}
\input{sections/evaluation}

\section{Conclusion}
\input{sections/conclusion}

\section{Replication data and source code}
\input{sections/replication}

\section*{Acknowledgements}
\input{sections/acknowledgements}

\section*{Bibliography}
\bibliographystyle{apalike}
\bibliography{main.bib}

\end{document}

%% file: sections/abstract.tex
\begin{abstract}


I present a novel approach to estimating the temporal and spatial patterns of violent conflict. I show how we can use highly temporally and spatially disaggregated data on conflict events in tandem with Gaussian processes to estimate temporospatial conflict trends. These trends can be studied to gain insight into conflict traps, diffusion and tempo-spatial conflict exposure in general; they can also be used to control for such phenomenons given other estimation tasks; lastly, the approach allow us to extrapolate the estimated tempo-spatial conflict patterns into future temporal units, thus facilitating powerful, stat-of-the-art, conflict forecasts. Importantly, these results are achieved via a relatively parsimonious framework using only one data source: past conflict patterns.\par

\end{abstract}
\pagebreak

%% file: sections/introduction.tex
In an effort to inhibit the onset and diffusion of violent conflicts, peace researchers have long sought to create reliable conflict forecasts \citep{singer1973peace}. Recently, researchers have started developing advanced computational frameworks, \emph{early-warning systems}, for predicting future conflicts \citep{hegre2019views}. In the short run, such systems will provide actors with valuable information and time to act, mitigating the fallout of conflicts \citep{Ward_Greenhill_Bakke_2010, perry_2013, hegre2017introduction, hegre2019views}. In the long run, such systems will generate theoretical insights into the mechanics of conflicts \citep{Schrodt_2014, chadefaux2017conflict, hegre2017evaluating, cederman2017Gurr}. Indeed, efforts to create early-warnings system have been called \enquote{conflict researchers' ultimate frontier} \citep[474]{cederman2017predicting}.\par

Most modern early-warning systems are compiled using individual components (or themes) each focusing on specific dimensions of conflict such as geography, demography, and socioeconomic factors \citep{ward2017lessons, hegre2019views}. Usually, at least one component represents the local and regional history of conflict -- i.e., the level of tempo-spatial conflict exposure a given location experiences at different points in time. Notably, such tempo-spatial components are consistently shown to be the most powerful predictors when forecasting conflict \citep{hegre2019views}. Turning to efforts which focus less on predictions and more on parameter estimation, tempo-spatial components are usually included to control for phenomena such as conflict traps and conflict diffusion \citep{beck1998taking, Hegre_Sambanis_2006, buhaug2008contagion, Goldstone_2010, schutte2011diffusion, crost2015conflict, hegre2017evaluating, bara_2017}. As such, the estimation and extrapolation of to tempo-spatial conflict exposure should be a central objective in both forecasting and estimation efforts.\par

In this article, I argue that, despite the importance of conflict exposure, we have previously relied too much on ad hoc specifications when constructing features and components meant to capture the tempo-spatial patterns of conflict exposure. One consequence of this is that we have not yet been able to effectively utilize the predictive potential of these patterns in our forecasting efforts. Another consequence is that past specifications often (implicitly) impose potentially imprudent assumptions onto these patterns, effectively preventing us from assessing the reach and consequences of temporal and spatial conflict exposure

The contribution of this article is a novel approach addressing this challenge. Importantly, the aim is not to create a new comprehensive early warning system, taking into account all relevant features pertaining to the demography, geography, political structures, socioeconomic factors, and so on. Nor is my approach meant to (or expected to) outperform existing state-of-the-art early warning systems. I focus exclusively on one component from one data source: tempo-spatial conflict exposure estimated and extrapolated using data on past conflict events. My objective is to show how we might realized the predictive power inherent in past conflict patterns while also producing insight regarding how far conflict exposure radiates through time and space.\par 

To this end, I use data on conflict events disaggregated at the monthly and sub-national level in tandem with \emph{Gaussian processes}. This allows me to estimate the underlying tempo-spatial patterns of conflict and further extrapolate these patterns into the future, thus generating an estimate of tempo-spatial conflict exposure. Intriguingly, this approach allows me to estimate both long and short-term conflict trends as well as spatial diffusion patterns. Furthermore, the approach also generates valid estimates of uncertainty pertaining to these trends. This allows me asses how far into future any generated forecast should be relied upon.\par

To assess my approach, I use the estimated tempo-spatial trends as features in a supervised machine learning ensemble to produce conflict forecasts at highly disaggregated spatial and temporal levels. In order to facilitate a comparative point of reference, I base my effort on the current state-of-the-art framework: ViEWS \citep{hegre2019views}. As such, I use the same data and the same temporal and geographical units, i.e., monthly PRIO-grid cells covering Africa. I use the same subset of the data for \emph{training}, \emph{validation}, and \emph{testing}, respectively. I use the same target and a similar machine learning ensemble for the final forecasting task. And finally, I used the same metrics to evaluate my approach. The only elements that differ substantially between this effort in \citet{hegre2019views} are the features I use; In this effort, these are all derived from data on past conflict patterns using Gaussian Processes. At no point do I incorporate any other features into my models.\par

With an $AP = 0.2704$ and an $AUC = 0.9318$ on a $36$ month forecasting window, my approach is able to credibly outperform ViEWS' comparable and thematically similar \enquote*{conflict history} component. Adding ViEWS social and natural geography components to their conflict history component lessens the gap, but my approach still prevails. Comparing my approach to a ViEWS model, which incorporates all its components, does not change this. Only ViEWS' final and best ensemble is able to marginally outperform my approach on a $36$ month forecast average. Intriguingly, though, my approach is able to outperform ViEWS' best ensemble during the first three months of the forecast. The prevalence of ViEWS best ensemble given the full test set of $36$ months should not be surprising: it is a far larger and more comprehensive framework with hundreds more features compared to the approach I present. What is surprising, however, is how well my rather parsimonious approach, which uses only past patterns, performs compared to a grand state-of-the-art early-warning system such as ViEWS.\par

Another attractive property of the approach is the fact that it generates useful uncertainty estimations. Thus I can asses how the uncertainty proliferates as I extrapolate the estimated patters forward into (test) time. Even more, I obtain heuristics regarding the temporal limit of my approach forecasting ability. Such insights carry both theoretical and practical relevance.\par

As such, my approach is a relatively parsimonious, yet powerful, framework capable of estimating and extrapolating sub-national patterns of tempo-spatial conflicts exposure. However, while I do illustrate the power of my approach by generating conflict forecasts and comparing these with ViEWS, I want to stress that the approach should still only be understood as one component: A component aimed at representing tempo-spatial conflict exposure. This component can be incorporated into more comprehensive early-warning-systems such as ViEWS \citep{hegre2019views}; it can be used as control variables for parameter-estimation and causal-identification; or it can be used to asses the expected tempo-spatial exposure generated by existing or hypothetical conflicts.\par

%% file: sections/theories.tex
By now, it has been firmly established that conflicts cluster in time and space, with \citet{crost2015conflict} listing no less than 20 published academic papers supporting this assertion \citep[15]{crost2015conflict}. As such, understanding how conflict diffuses through these dimensions has the potential to enhance our conflict forecasts greatly. To motivate my specific approach, I here supply a short review of the literature pertaining to temporal and spatial conflict patterns.\par

\subsection{Temporal conflict patterns}

We know that conflict in one geographic region is dependent on the local conflict history \citep[1263-1264]{beck1998taking}. This phenomenon is known as a \emph{conflict trap} \citep{hegre2017evaluating}. Once some geographic region gets embroiled in conflict, a roster of derived effects starts to inhibit the conclusion of said conflict as well as increasing the risk of future conflicts. The development of political mobilisation, military socialisation and militarisation of local authorities lingers and increases the risk of prolonged/recurring conflicts \citep{wood2008social}. Conflicts can strengthen the influence of the military and award power to individuals profiting from the violence \citep{hegre2017evaluating}. Such \enquote*{spoilers} might obstruct peace processes to maintain or gain influence \citep{stedman1997spoiler}. The fragmentation of the local political economy, the disintegration of social networks and the polarisation of social identities might also make reconciliation difficult \citep{wood2008social}. Veterans can be difficult to reintegrate into society \citep{humphreys2007demobilization}, firearms might circulate in large numbers \citep{lock1997armed}, and inter-group grievances can linger and proliferate \citep{hegre2017evaluating}. Capital and skilled labour might flee and hesitate to return even if the conflict is concluded \citep{collier1999economic}. Infrastructure is destroyed, debt incurred \citep{slantchev2012borrowed}, and trade disrupted \citep{bayer2004effects}. In turn, this will lower wealth and growth, thus making a region more prone to renewed conflict by lowering the state capacity \citep{Fearon_Laitin_2003} and/or by lowering the opportunity cost of rebelling \citep{Collier_Hoeffler_2004}. Naturally, conflicts might end at some point -- when the parties deplete their resources, manpower or political capital \citep{mason1999win}, one side wins or an enduring settlement is made \citep{toft2010ending} -- but achieving such a conclusion is made difficult given the inertia induced by the presented factors.\par

These theoretical expectations are supported by empirical findings. For instance, it has been found that the probability of conflict reoccurring/continuing in a specific country can increase up to $20\%$ in the first year following a conflict event. Hereafter, the impact of past events decreases each year but might still be observed $25$ years later \citep[249]{hegre2017evaluating}. Furthermore, the severity of a conflict trap appears to be a function of the magnitude of conflict: more conflict, longer trap \citep[255]{hegre2017evaluating}. It follows that an increase in magnitude might imply that a conflict trap is tightening, while a decrease in conflict magnitude might signal a loosening of a trap \citep[158-159]{hegre2017evaluating}. Lastly, it has been shown that temporal conflict patterns can be divided into short- and long-term trends \citep{hegre2021can}. In sum, once a conflict has started, we should expect it to extend into the future via some relatively torpid temporal pattern. The level of conflict can be increasing or declining, but future patterns should be influenced by past patterns.\par

\subsection{Spatial conflict patterns}
 
Conflicts can also diffuse through space \citep[246]{hegre2017evaluating}. Indeed, the emerging consensus is that the clustering of conflict is often a direct product of diffusion rather than simply a byproduct of other clustered factors \citep{buhaug2008contagion, schutte2011diffusion, crost2015conflict, hegre2017evaluating, bara_2017}. Many conflict-facilitating factors extending through time also diffuse through space. Spoilers might seek to expand their influence. Grievances, fighters, veterans, and firearms can spread from the original conflict to nearby areas. Capital, skilled labour and trade might not just leave conflict areas but also the general vicinity of conflict. Thus, risk-inducing factors such as poverty, deprivation and insufficient state capacity might spread as a product of conflict itself. Furthermore, influence and support from trans-border ethnic kin might increase conflict potential \citep{Cederman_Gleditsch_Buhaug_2013}, thus impeding resolution.\par

Using disaggregated data, it has been found that patterns of internal conflict are characterised by distinctive spatial patterns comparable across cases: When conflict engulfs one location it is likely to expand outwards from said location over time \citep[151]{schutte2011diffusion}, often disregarding both structural factors and administrative boundaries \citep[442-443]{ol2010afghanistan}. When employing disaggregated geographic units-of-analysis, this pattern of conflict can, at its most fundamental level, be seen as exhibiting bell-curve-like (or Gaussian) properties. That is, the risk of conflict diffusing to a given location is influenced by said location's distance to any active conflict, along with the magnitude of this conflict.\par

\subsection{Past operationalisations of conflict patterns}

While there exists a relatively broad agreement regarding the theoretical mechanisms just presented, little consensus has been reached regarding how these mechanisms should be modelled. Starting with the temporal patterns of conflict \citet{Hegre_Sambanis_2006} use a linear decay function counting years since last peace, while \citet{Collier_Hoeffler_2004} and \citet{hegre2019views} use a linear decay function counting years since last conflict. \citet{perry_2013} suggests an index akin to a linear decay function since the last conflict, but scaled by the magnitude of the specific conflict. All these operationalisations aim to capture the effect of a conflict trap.\par

The danger is that such operationalisations rely too much on manual specification, implicitly leading to potential imprudent assumptions. For instance, \citet{hegre2019views} specifies the half-life parameter of their decay function to $12$ months, yet provides little explicit justification for this decision. This specification is likely backed by qualitative knowledge and might be reasonable, but needing such ad hoc decisions is still a weakness of decay functions \citep[501]{Gelman_2013}. The issue intensifies when using more elaborate features such as the index proposed by \citet{perry_2013}. While the incorporation of past magnitude is highly warranted, the specifications of the scale parameter appear ad hoc. Again, the specifications were likely chosen on grounds of good reason, but they are still assumed. If such assumptions lead to misspecification, it will impact prediction power negatively.\par

The solution is to employ methods capable of estimating functional forms and parameters directly from the data. And where (hyper) parameters cannot be directly estimated, they should be chosen through out-of-sample validation. This approach will minimize the number of assumptions and ad hoc decisions needed, which in turn will ensure more fitting specifications and lead to both increased predictive power and substantial insights.\par

Turning to the spatial patterns of conflict, the challenges are rather similar. A binary feature ($\in\{0,1\}$) has often been used to indicate whether some predefined number of neighbouring regions are experiencing conflict. For instance, \citet{ol2010afghanistan}, \citet{weidmann_ward_2010predicting} and \citet{hegre2019views} all include binary features for conflict in $1^{st}$ order neighbouring units. However, given the relatively small unit-of-analysis employed, this is likely insufficient. When employing disaggregated geographical units $2^{nd}$, $3^{th}$, $4^{th}$ and potential $n^{th}$ order neighbours must be considered. Yet, even if we included features for the $2^{nd}$ nearest conflict, the $3^{th}$ nearest conflict etc., and scale the distance by some factor related to the magnitude of conflict, we still do not know if these conflicts surround and engulf the observation of interest or instead cluster distinctly beside it. The geographic pattern of conflict might matter, as could the magnitude of conflict as well as the magnitude of conflicts in $1^{st}$, $2^{nd}$, and $n^{th}$ order neighbours. If we do not manage to model the spatial patterns of conflict accordingly we will lose predictive power to misspecification and leave unnoticed theoretically interesting insights.\par

The solution is to generate functions capable of capturing the underlying spatial patterns directly from the data. And where estimation is infeasible, out-of-sample validation should be employed. As such, we avoid ad hoc operationalisation.\par

To summarise: In regard to the temporal patterns, we need an approach which allows the incorporation of information drawn from all past data available. The magnitude of past conflicts should be allowed to influence the probability of future conflicts with some decreasing influence as a function of time. Furthermore, this deterioration rate should be estimated through the data itself, not assumed by the researcher. Given the spatial patterns, we need an approach capable of including signals from all relevant events, taking into account both the magnitude and distance of all relevant conflicts. We also want to distinguish between different patterns of conflict. For instance, actual encirclement should likely lead to greater exposure compared to mere adjacency. Importantly, researchers should not decide how close a conflict has to be in order to be included, nor the deterioration rate of influence. Such specifications should instead be estimated directly from the available data. As I will demonstrate going forward, these criteria can be accommodated using Gaussian processes.\par

%% file: sections/data.tex
To estimate the patterns of conflict, I need data that is highly disaggregated at both the temporal and spatial levels. I also want data used by similar forecasting efforts in order to ensure a valid basis of comparison. Using the replication data available from \citet{hegre2019views} fulfills these criteria.\par

A thorough presentation of the data can be found in the online appendix. In short, the units-of-analysis are monthly \emph{PRIO-grid cells} (squares roughly measuring $50km\times50km$ at the equator \citep[367]{Tollefsen_2012}). \citet{hegre2019views} use a geographical subset covering Africa and focuses on \emph{state-based conflict} ($sb$). For comparability,y I do the same. \citet{hegre2019views} use the UCDP to obtain data on conflict events, specifically a feature pertaining to fatality counts (denoted $best$). All data used from here on comes from this single measure $best_{sb}$. All features will be derived from this measure, and all targets will be based on it -- past patterns will predict future patterns. I take the logarithm of $best_{sb}$ to create the measure \emph{conflict magnitude} ($cm$). Justification for this transformation can be found in the online appendix. To create the target for the final classification tas,k I follow \citet{hegre2019views} and apply a binary transformation of the original measure $best_{sb} > 0 := 1$, denoting whether or not some unit-of-analysis experience conflict or not.\par

To evaluate the performance of my approach, I use \emph{out-of-sample} prediction \citep{Ward_Greenhill_Bakke_2010, Goldstone_2010, hegre2019views}. Again, I model my approach directly after \citet{hegre2019views}. \citet{hegre2019views} use data from $1990$ through $2011$ for training while data from $2012$ through $2014$ is used for out-of-sample validation. During the final evaluation, the training and validation sets are combined into a larger set ($1990$ through $2014$) and the last $36$ months of data ($2015$ through $2017$) are used as a \emph{hold-out test set} for out-of-sample evaluation \citep[163]{hegre2019views}.\par

The challenge now is to effectively estimate and extrapolate relevant patterns from this data. Given the theoretical and empirical insights presented above, I propose we use Gaussian processes to address this challenge.\par

%% file: sections/gps.tex
The main takeaway point from the sections above is that, given theory and past empirics, we can safely assume that there exists a latent, relatively inert, pattern of conflict exposure, both temporal and spatial. Such patterns can be approximated by a combination of smooth\footnote{By smooth I here mean continues and differentiable.} functions (plus error terms). It is trivial to fit such a model to any data, the challenge is to do so without imposing too many imprudent assumptions and to do so without overfitting the model to the observed data. As I will demonstrate below, this is where Gaussian Processes can prove a powerful tool.\par

After a brief introduction to Gaussian Processes, I will proceed to my implementation and estimate two types of conflict exposure. The first estimate is one of the temporal exposure, which only takes into account the conflict history of the given grid cell. I denote this pattern \emph{temporal conflict exposure} ($TCE$). The second is an estimate that also takes into account the conflict history of vicinal grid cells. I denote this pattern \emph{tempo-spatial conflict exposure} ($TSCE$).\par

The use of Gaussian processes allows me to estimate parameters and functions that are usually assumed. However, there are still decisions to make regarding various specifications and hyperparameters. To avoid arbitrary or imprudent choices, I use out-of-sample validation to find the best-performing specifications. This approach is routinely used when evaluating forecasting frameworks and is also applied in \citet{hegre2019views}. Thus, the specific implementation presented below is that which displayed the most predictive power during validation.\par

\subsection{Estimating temporal conflict exposure}

To illustrate how Gaussian processes work, I start with the temporal dimension of conflict. I will refer to the combined months of a given grid cell as the \emph{timeline} of said cell. Taking a single time-line pertaining to some geographic cell, we can understand said time-line as a continuous function $y = f(x) + \epsilon$. Here $f$ is a smooth function, $\epsilon$ is a noise term, $x$ is a vector of months, and $y$ is a vector of corresponding, observed, conflict magnitudes.\par

We can use Gaussian processes to fit and estimate distributions of (smooth) functions from which we can sample arbitrarily many $f$ \citep[13-15]{williams2006gaussian}. The mean of such a sample can be used as a \emph{maximum likelihood} estimate, $\mu$, representing the most likely form of the \enquote*{true} latent function. Naturally, we can also estimate the standard deviation, $\sigma$, of such a sample, which we can use to quantify how certain (or uncertain) we are regarding $\mu$. 

So, the example given by equation \ref{eq:y} and \ref{eq:f} denotes that $y$ is given by an error terms $\epsilon$, and a function $f(x)$ which in turn is drawn from a Gaussian process (here denoted $\mathcal{GP}$). The Gaussian process, in turn, is given by a \emph{mean function} ($m$) and a \emph{covariance function} ($k$). The $\epsilon$ is included and estimated to avoid \emph{overfitting} to noise \citep{williams2006gaussian, Mcelreath_2018}.\par

\[
y = f(x) + \epsilon \tag{1} \label{eq:y}
\]

\[
f(x) \sim \mathcal{GP}(m(x),k(x,x')) \tag{2} \label{eq:f}
\]

The mean and covariance functions determine the properties of the functions we can sample from the Gaussian process. E.g., how volatile and flexible they will behave. This is central since one attribute of smooth functions is that vicinal values are constrained to be relatively similar. How far two vicinal values can diverge depends on the flexibility of the function and the distance between the two values. Close vicinity and inflexible functions lead to very constrained and similar values, while larger distance and more flexibility allows more freedom for the values to vary.\par

To be more specific, the mean function $m$ denotes the value $f$ takes in the absence of any relevant information. For instance, we can always add new months ($x_{new}$) to our model and generate corresponding expected values $y_{new}$. Naturally, any $y_{new}$ close to observed values would be expected to be relatively similar to the observed values, and the function would be expected to follow the trajectory from the observed past into the unobserved future. However, as we move further into the future, without any new observed conflicts, the signal will at some point cease to inform our model. Here our maximum likelihood estimate $\mu$ defaults to the mean function $m$ \citep[3]{williams2006gaussian}. Often, $m$ is set to the constant zero \citep[28-29]{williams2006gaussian}. Since I do not extrapolate any forecasts far enough for the mean function to become overly consequential, I keep to this convention and denote this mean function $m_{0}$.\par 

The covariance function $k$ is far more consequential as it describes the similarity of $y$ between all observations $x$ and $x'$. As such, it is this function which effectuates the assumption that two observations with similar values on $x$ likely also have similar values on $y$ \citep[79]{williams2006gaussian}. Given a specific timeline, this implies that units close in time are expected to experience relatively similar amounts of temporal conflict exposure. Given a set of contemporary grid cells, it implies that units close in space should experience relatively similar amounts of spatial conflict exposure. And lastly, given a set of timelines, it implies that units close in both time and space should experience relatively similar amounts of tempo-spatial conflict exposure.\par

I employ two different covariance functions going forward: The \emph{squared exponential} ($k_{SE}$) covariance function and a Matérn covariance function ($k_{Matern32}$). The squared exponential covariance function is widely applied since the only major assumption attached to it is that we are estimating smooth functions \citep[84]{williams2006gaussian}. As such, it is often used to model temporal trends \citep[119]{williams2006gaussian}. The Matérn covariance function is closely related to the squared exponential covariance function \citep[84-85]{williams2006gaussian} but can be more flexible \citep{genton2001classes, minasny2005matern}. It has a rich history in the geostatistical literature \citep[1408]{melkumyan2011multi} where it is used to model spatial covariance \citep[193]{minasny2005matern}. However, it can also be used to model temporal trends where it allows for rougher functional forms. The covariance function $k_{SE}$ is given in equation \ref{eq:k_se} and $k_{Matern32}$ in equation \ref{eq:k_mat}.\par

\[
k_{SE}(x,x') = \eta^2 exp\left(-\frac{|x-x'|^2}{2\ell^2}\right) \tag{3} \label{eq:k_se}
\]

\[
k_{Matern32}(x,x') = \eta^2 \left(1+ \frac{\sqrt{3|x-x'|^2}}{\ell} \right) exp\left(-\frac{\sqrt{3|x-x'|^2}}{\ell} \right) \tag{4} \label{eq:k_mat}
\]

What stands out in both $k$'s is the hyperparameters $\ell$ and $\eta$. These are scale parameters which can be estimated from the data and determine the volatility and flexibility of the estimated functions\footnote{These estimates are obtained by maximising the \emph{marginal likelihood} of the given model \citep[114-115]{williams2006gaussian}. Readers can consult \citet{williams2006gaussian} section 5.4.1 for an overview}. The $\eta$ is denoted \emph{amplitude} and roughly determines the average distance the functions vary from their means \citep[502]{Gelman_2013}. For our purpose, however, $\ell$ will be the most interesting hyperparameter going forward.\par 

The $\ell$ is denoted the \emph{lengthscale} \citep[501-502]{Gelman_2013} and can be interpreted as the \enquote*{distance} we have to move across the $x-axis$ before the value of $y$ can change substantially \citep[14-15]{williams2006gaussian}. Together with the specific covariance function used, it is really this hyperparameter that determines the flexibility of the sampled functions. Here this translate to the reach of conflict exposure through time and space. This is especially important when forecasting out-of-sample into future months: A large $\ell$ equates a rigid function while a small $\ell$ equates a more flexible function. Naturally rigid functions will be more constrained by past observed data compared to flexible functions. This also means past observed data will inform any extrapolations further into the unobserved future, given a large $\ell$. Conversely, when we sample more flexible functions, the signal from the past data will quickly wane. Indeed, we can use $\ell$ as an heuristic regarding how far into the future we should consult our forecasts: Beyond the last observed training month + $\ell$ the data provides little signal, uncertainty proliferates (i.e. $\sigma$ increases) and the mean function $m_{0}$ begins to dominate.\par 

It should be clear by now that the observed data is a central part of $m$ and $k$ and thus any Gaussian process. This differs from e.g. linear models where data is summarized by a finite number of parameters. Gaussian processes are said to be \emph{non-parametric} models, but you could see $m(x)$ and $k(x,x')$ as parameters. Given this view, we have as many parameters as we have data points. And since we, in theory, can have infinite data points, we can also have infinite parameters - which is one definition of non-parametric models \citep[166-167]{williams2006gaussian}.\par

Now, I do not use all timelines to estimate the hyperparameters $\eta$ and $\ell$. The reason is that most grid cells do not experience any conflicts throughout their timeline. Including such \enquote*{flat-lines} does not tell us anything about the patterns of conflict and only serves to bias the estimated hyper-parameters by leading to overly large $\ell$s. Via the validation step, I find that including timelines that have, at some point, experienced at least eight months of conflict during one year works very well. Notably, there is a range of similar criteria where comparable results are achieved, indicating that we simply need some amount of relatively coherent conflict to estimate the underlying patterns.\par

Through the validation step, I also found that a two-trend model outperformed all other tested specifications. This aligns with theory and past empirics, which have found both short- and long-term trends \citep{hegre2017evaluating, hegre2021can}. I found that using $k_{SE}$ for the long-term trend and $k_{Matern32}$ for the short-term trend proved the most powerful specification. This is not overly surprising given the insights presented regarding conflict traps: We would expect some long underlying smooth trend/trap accompanied by a rougher short-term trend. Implementing two trends is simply done by letting $f$ be the sum of two sub-functions\footnote{Determining which function becomes the long term trend and which becomes the short term trend is simply a matter of setting slightly informative priors on the hyper-parameters $\ell_{TCEshort}$ and $\ell_{TCElong}$}.\par 

The formal expressions are presented in \ref{eq:f_tce}, \ref{eq:tcelongshort}, \ref{eq:tce_long} and \ref{eq:tce_short}. I add a subscript $t$ here to highlight that the $x_{t}$ is a vector of temporal units (months). I also add the subscript $cm$ to denote that $y_{cm}$ is conflict magnitude (conflict fatalities logged). The parameters are subscribed with $TCE$ to denote that these pertain to the estimated patterns of temporal conflict exposure. The parameters are furthermore given the subscript $long$ if they pertain to the long-term trend and $short$ if they pertain to the short-term trends.\par 

\[
y_{cm} = f_{TCE}(x_{t}) + \epsilon \tag{5} \label{eq:f_tce}
\]

\[
f_{TCE}(x_{t}) = f_{TCElong}(x_{t}) + f_{TCEshort}(x_{t}) \tag{6} \label{eq:tcelongshort}
\]

\[
f_{TCElong}(x_{t}) \sim \mathcal{GP}(m_{0}(x_{t}),k_{SE}(x_{t},x_{t}')) \tag{7} \label{eq:tce_long}
\]

\[
f_{TCEshort}(x_{t}) \sim \mathcal{GP}(m_{0}(x_{t}),k_{Matern32}(x_{t},x_{t}')) \tag{8} \label{eq:tce_short}
\]

After using the validation step to find the final form, the training and validation sets are combined, and the hyperparameters are re-estimated. These estimated hyperparameters are presented in \autoref{tce_hp} and are virtually identical to the ones estimated during the validation step.\par

\begin{table}[!htb]
\centering
\caption{\textbf{hyper-parameters}\\Temporal conflict exposure}\label{tce_hp}
	\begin{tabular}{m{4cm} m{9cm}}
	\toprule
                            &  \thead{Point estimate\\(MAP)}    \\
	\midrule
	$\ell_{TCEshort}$        & \thead{4.09}                       \\
	$\eta_{TCEshort}$        & \thead{0.32}                       \\
	$\ell_{TCElong}$         & \thead{122.38}                     \\
    $\eta_{TCElong}$         & \thead{0.50}                       \\
    $\epsilon_{TCE}$         & \thead{0.8}                        \\
    &\\

    \bottomrule
	\end{tabular}
\begin{tablenotes}
  \item Note: The table contains the estimated hyperparameters pertaining to $f_{TCE}$. For computational efficiency, I only estimate the \enquote*{Maximum A Posterior} ($MAP$), which is simply a point estimate of the mode given the full posterior distribution \citep[42]{Mcelreath_2018}. Initial test indicated that estimating a full posterior distribution did not change these hyperparameters and that the credibility interval around them where reasonably tight.
\end{tablenotes}
\end{table}

The fact that $\ell_{TCEshort}$ is estimated to be $4.09$ tells us that little to no signal from the short-term trend reaches beyond four months into the $ 36$-month long test set. $\ell_{TCElong}$ is estimated to be $122.38$ meaning that signals from the long term trend radiates $10$ years in the future -- easily covering all $36$ test months. While extrapolating conflict patterns $10$ years ahead might sound overconfident, it has been shown that it is possible to generate informative risk assessments over such long temporal spans \citep{hegre2017evaluating, hegre2021can}.\par

\begin{figure}[!htb]
	\centering
	\includegraphics[scale=0.6]{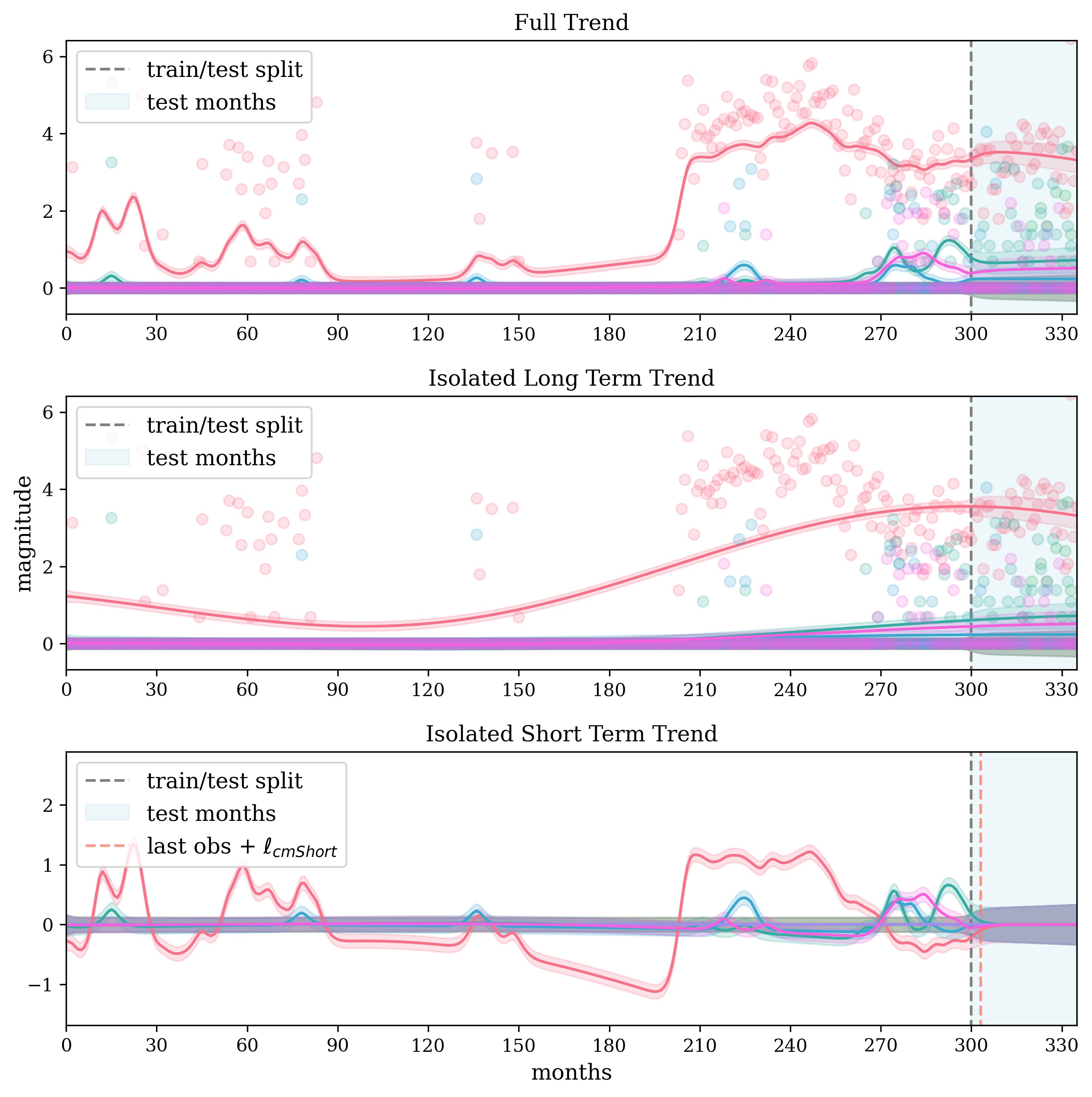}
    \caption{\footnotesize{An illustrative sample of eight adjacent timelines. From the top the full trend $\mu_{TCE}$, then $\mu_{TCElong}$ and $\mu_{TCEshort}$. The scatter points show the observed conflict magnitude for each timeline. The functions are estimated using the 300 months in the training set and then extrapolated into the test set. The gray dashed line indicates the transition from in-sample (training set) to out-of-sample (test set). Thus, while both the observed conflict magnitudes from the train and test months are plotted, only the observations from the training months have been used to estimate the functions.}}\label{tce_sampel_eks}
\end{figure}

I now extrapolate the functions into the future months by introducing $x_{tnew}$. I.e., the months in the test set. Note that I do not use any data on conflict magnitude from the test set to do this -- only data from the train set, the hyperparameters estimated, and the specifications presented above. I have drawn and plotted eight time-lines in \autoref{tce_sampel_eks} below to illustrate these strictly temporal trends. First, the full trend $\mu_{TCE}$, then the isolated long-term trend $\mu_{TCElong}$, and then the isolated short trend $\mu_{TCEshort}$. The gray line indicates the transition from training to test month. Since $\ell_{TCEshort} < 36$, the plot depicting the short-term trend also includes a red line indicating the last observed month plus $\ell_{TCEshort}$.\par

The consequence of the rather short $\ell_{TCEshort}$ is that $\mu_{TCEshort}$ relatively quickly defaults to the mean function $m_{0}$, leaving us with the signal provided by $\mu_{TCElong}$. While this does not necessarily render $\mu_{TCEshort}$ irrelevant for predictive purposes it does mean that $\mu_{TCElong}$ will be more important given the $36$ month forecasting window here employed.\par

Since the estimated $\mu$'s effectively represent continuous functions, I can readily extract more relevant information. Specifically how much the temporal conflict exposure (and later tempo-spatial conflict exposure) is increasing or decreasing a given month (slope: $\mu'$), whether the level of exposure is accelerating or decelerating (acceleration: $\mu''$) and the total past exposure to conflicts with a given time-line (cumulative sum: $M$). Now, the actual outputs are the values generated by the functions, not any formal expressions. Thus $\mu', \mu''$ are derived through gradient approximation and the $M$ through cumulative summation. I apply this step to $\mu_{TCE}$, $\mu_{TCElong}$ and $\mu_{TCEshort}$, thus deriving $\mu'_{TCE}, \mu''_{TCE}, M_{TCE}, \mu'_{TCElong}, \mu''_{TCElong}, M_{TCElong}, \mu'_{TCEshort}, \mu''_{TCEshort}$ and $M_{TCEshort}$. The values taken directly from the functions $\mu_{TCE}*$ are likely relevant since the magnitude and order of past conflicts matter. The derived features $\mu'_{TCE}*$ and $\mu''_{TCE}*$ might be relevant since we know that the relative change in conflict can signal different stages in a conflict trap. The cumulative sums $M_{TCE}*$ might be relevant since conflict traps can span decades.\par

Revisiting the guiding criteria pertaining to the temporal dimension, we see that this approach fulfils them all. It allows for the incorporation of information drawn from all past (included) months with decreasing influence. Magnitude matters, and the rate at which past months influence the forecasts is estimated through the data itself. Furthermore, no arbitrary \enquote*{splines} or \enquote*{knots} needs defining, which is one reason it has been recommended as a substitute to linear decay functions \citep[501]{Gelman_2013}. Lastly, I am also able to easily identify and estimate different trends and asses the potential reach of conflict exposure through time.\par 

\subsection{Estimating tempo-spatial conflict exposure}

The patterns estimated above only take into account a given cell's past history of conflict, with no inclusion of spatial information. To amend this I now create features pertaining to \emph{tempo-spatial conflict exposure} ($TSCE$).\par

We can estimate Gaussian processes over 2 spatial dimensions (longitude and latitude) just as we did with 1 temporal dimension (months). In theory, we could estimate a $3D$ function over time and space simultaneously. In practice, this is prohibitively \enquote*{expensive} as Gaussian Processes do not scale well to higher dimensions given large data sets \citep[171]{williams2006gaussian}. My solution is to first estimate a $2D$ function over the geographical space of each individual month in the training set\footnote{First, leaving the validation set out, and later incorporating it for the final evaluation against the test set}. Subsequentl,y I estimate $1D$ functions across the temporal dimension of these 2d functions. This second step is analogous to the estimation of $TCE$ above, but now using the estimated spatial exposure as a target instead of the conflict magnitude.\par

To estimate 2D functions over the geographical spaces, I use the spatial units $X_{l,l}$. This is simply a $n \times 2$ matrix for each month in the dataset, with one column containing latitude coordinates for each grid cell and one containing corresponding longitude coordinates (centroids of the grid cells). The target here is the same as before, conflict magnitude $y_{cm}$. Note, however, that while the hyperparameters are estimated using all months in the training set, I have not yet specified any explicit temporal connection through time. As such, I subscript the estimated function with $SCE$ for \emph{spatial conflict exposure}. The formal expressions given in equation \ref{eq:f_sce} and \ref{eq:sce} were chosen via out-of-sample validation.\par

\[
y_{cm} = f_{SCE}(X_{l,l}) + \epsilon \tag{8} \label{eq:f_sce}
\]

\[
f_{SCE}(X_{l,l}) \sim \mathcal{GP}(m_{0}(X_{l,l}),k_{Matern32}(X_{l,l},X_{l,l}')) \tag{9} \label{eq:sce}
\]

The covariance function $k_{Matern32}$ turned out to perform best during validation, which is unsurprising given its geostatistical origin. The choice of mean function $m_{0}$ is largely inconsequential here, as I do not extrapolate into any geographical space where I do not have data.\par 

Again, I use the validation step to find an optimal subset in order to avoid estimating overly long length scales\footnote{the computational cost of this estimation also dictates the use of a sparse implementation, focused on informative cells \citep[171]{williams2006gaussian}}. The validation step led me to use the $60$ cells experiencing the most conflict each given month. As there are rarely $60$ cells experiencing conflict, this leads to a varying number of non-conflict cells being included each month. Again, a range of similar criteria around this specification led to similar performances. The estimated hyper-parameters can be seen in \autoref{sce_hp}\par 

\begin{table}[!htb]
\centering
\caption{\textbf{hyper-parameters}\\Spatial conflict exposure}\label{sce_hp}
	\begin{tabular}{m{4cm} m{9cm}}
	\toprule
                        &  \thead{Point estimate\\(MAP)}  \\
	\midrule
	$\ell_{sce}$        & \thead{0.72}                     \\
	$\eta_{sce}$        & \thead{0.027}                    \\
    $\epsilon_{sce}$    & \thead{0.11}                     \\
    &\\

    \bottomrule
	\end{tabular}
\begin{tablenotes}
\item The table contains the estimated hyperparameters pertaining to $f_{SCE}$. For computational efficiency, I only estimate the $MAP$'s.
\end{tablenotes}
\end{table}

\begin{figure}[!htb]
	\centering
	\includegraphics[scale=0.6]{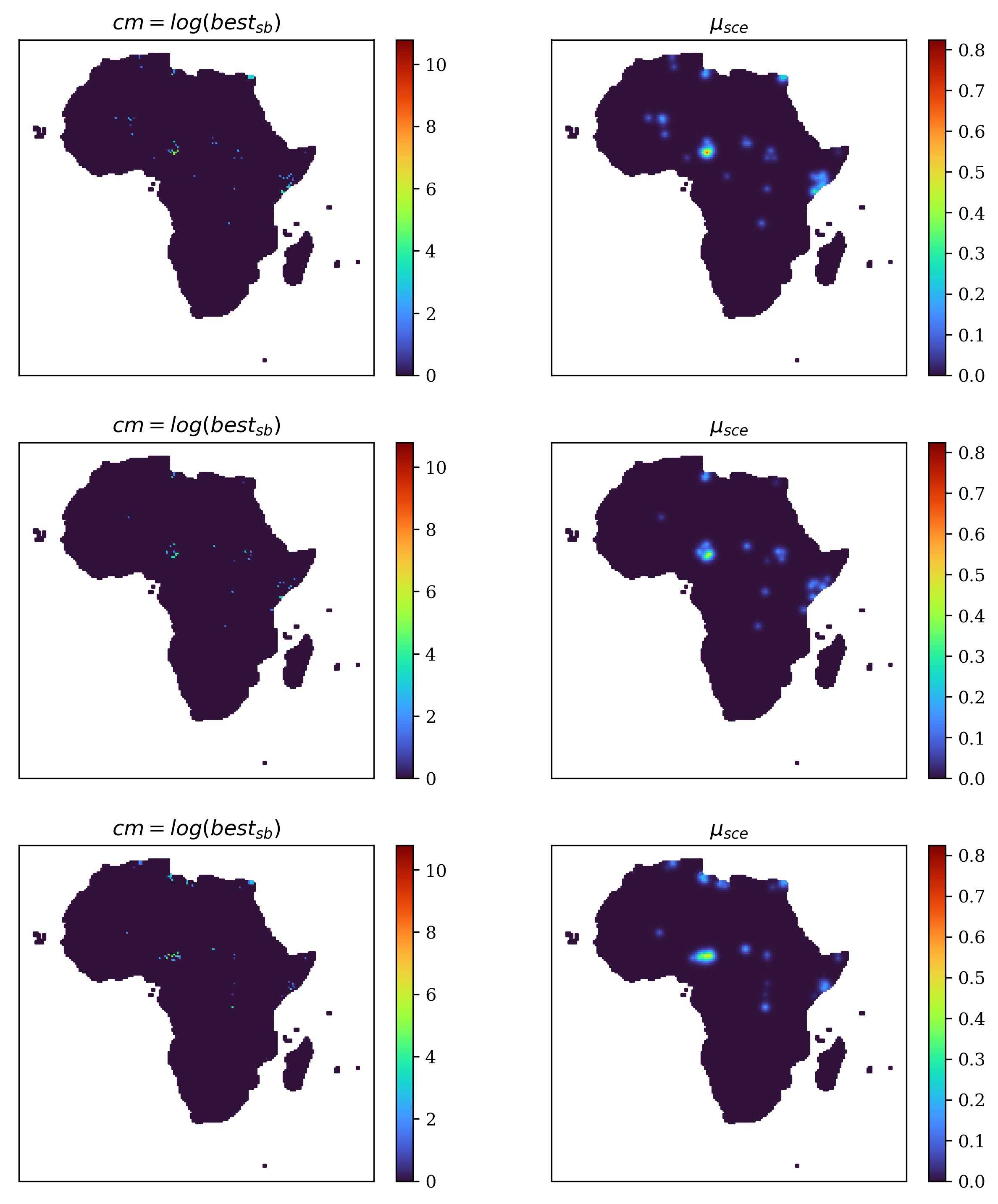}
    \caption{\footnotesize{Going forward in time from top to bottom, this is the last three months in the combined training/validation set  (October, November, and December $2014$). On the left we see the actual observed conflict fatalities (logged) and to the right the estimated spatial conflict exposure $\mu_{SCE}$.}}\label{sce_sample_eks}
\end{figure}

With each cell being $0.5$ decimal degrees and $\ell_{SCE} = 0.72$ even low-level conflicts will radiate into nearby cells. Importantly, the reach is a function of magnitude and thus higher levels of conflict will lead to higher levels of exposure and so greater potential of diffusion. This relationship becomes more tangible in \autoref{sce_sample_eks} where I have plotted the last three months of the training set.\par

In \autoref{sce_sample_eks} we see how higher conflict magnitude leads to greater reach. Thus, the \enquote*{rate of declining influence} is estimated directly from the data. Another desired criterion we see fulfilled pertains to adjacency vs. encirclement. Looking at northeastern Nigeria, southern Somalia, and the border between Sudan and South Sudan, we can observe that encircled cells experience higher spatial conflict exposure compared to cells not encircled. In short, each cell shares information regarding its magnitude of conflict with all cells in the relevant vicinity.\par

I finalize my \enquote*{pseudo-$3D$ approach} by estimating a $1D$ temporal Gaussian process over $\mu_{SCE}$. As noted, this is equivalent to the procedure used earlier to estimate and extrapolate $\mu_{TCE}$, but instead of using conflict magnitude $y_{cm}$ as my target, I now use the estimated spatial exposure $\mu_{SCE}$ as the target. As such, the function is again estimated over the temporal units (months). I denote this function \enquote*{tempo-spatial conflict exposure} ($TSCE$) since it takes into account the history of both local and regional conflict exposure. As seen in equation \ref{eq:f_tsce}, \ref{eq:tscelongshort}, \ref{eq:tsce_long}, and \ref{eq:tsce_short}, the validation step led me to use the same formulation as used to estimate $f_{TCE}$.\par 

\[
\mu_{SCE} = f_{TSCE}(x_{t}) + \epsilon \tag{10} \label{eq:f_tsce}
\]

\[
f_{TSCE}(x_{t}) = f_{TSCElong}(x_{t}) + f_{TSCEshort}(x_{t}) + \epsilon \tag{11} \label{eq:tscelongshort}
\]

\[
f_{TSCElong}(x_{t}) \sim \mathcal{GP}(m_{0}(x_{t}),k_{SE}(x_{t} ,x_{t}')) \tag{12} \label{eq:tsce_long}
\]

\[
f_{TSCEshort}(x_{t}) \sim \mathcal{GP}(m_{0}(x_{t}),k_{Matern32}(x_{t},x_{t}')) \tag{13} \label{eq:tsce_short}
\]

The derived features $\mu_{TSCE}$, $\mu_{TSCELong}$, and $\mu_{TSCEShort}$ should not be understood as strictly spatial measures, only considering signals from nearby conflicts. These features also include signals regarding the temporal history of each individual cell and all vicinal cells. The simplest way to understand these features is to see them as similar to $\mu_{TCElong}$ and $\mu_{TCEshort}$ but highly influenced by the surrounding temporal-spatial patterns of exposure. In Bayesian terms, we can think of this as a rather influential prior using relevant temporal-spatial information. In Frequentist terms, we can think of it as regularization towards some local temporal-spatial mean.\par

As before, the training and validation sets are combined after the final form has been found and the hyperparameters re-estimated. These hyper-parameters are reported in \autoref{tsce_hp}\par

\begin{table}[!htb]
\centering
\caption{\textbf{hyper-parameters}\\Tempo-spatial conflict exposure}\label{tsce_hp}
	\begin{tabular}{m{4cm} m{9cm}}
	\toprule
                            &  \thead{Point estimate\\(MAP)}  \\
	\midrule
	$\ell_{TSCEshort}$        & \thead{7.17}                    \\
	$\eta_{TSCEshort}$        & \thead{0.04}                    \\
	$\ell_{TSCElong}$         & \thead{74.72}                   \\
    $\eta_{TSCElong}$         & \thead{0.08}                    \\
    $\epsilon_{TSCE}$         & \thead{0.06}                    \\
    &\\

    \bottomrule
	\end{tabular}
\begin{tablenotes}
\item The table contains the estimated hyperparameters pertaining to $f_{TSCE}$. For computational efficiency, I only estimate the $MAP$.
\end{tablenotes}
\end{table}

With an $\ell_{TSCEshort} = 7.17$ the signal from this short-term trend will reach further into the forecast than $\ell_{TCEshort}$. Conversely, with $\ell_{TSCElong} = 74.72$, the signal from this long-term trend will have a shorter reach than $\ell_{TCElong}$.\par

\begin{figure}[!htb]
	\centering
	\includegraphics[scale=0.6]{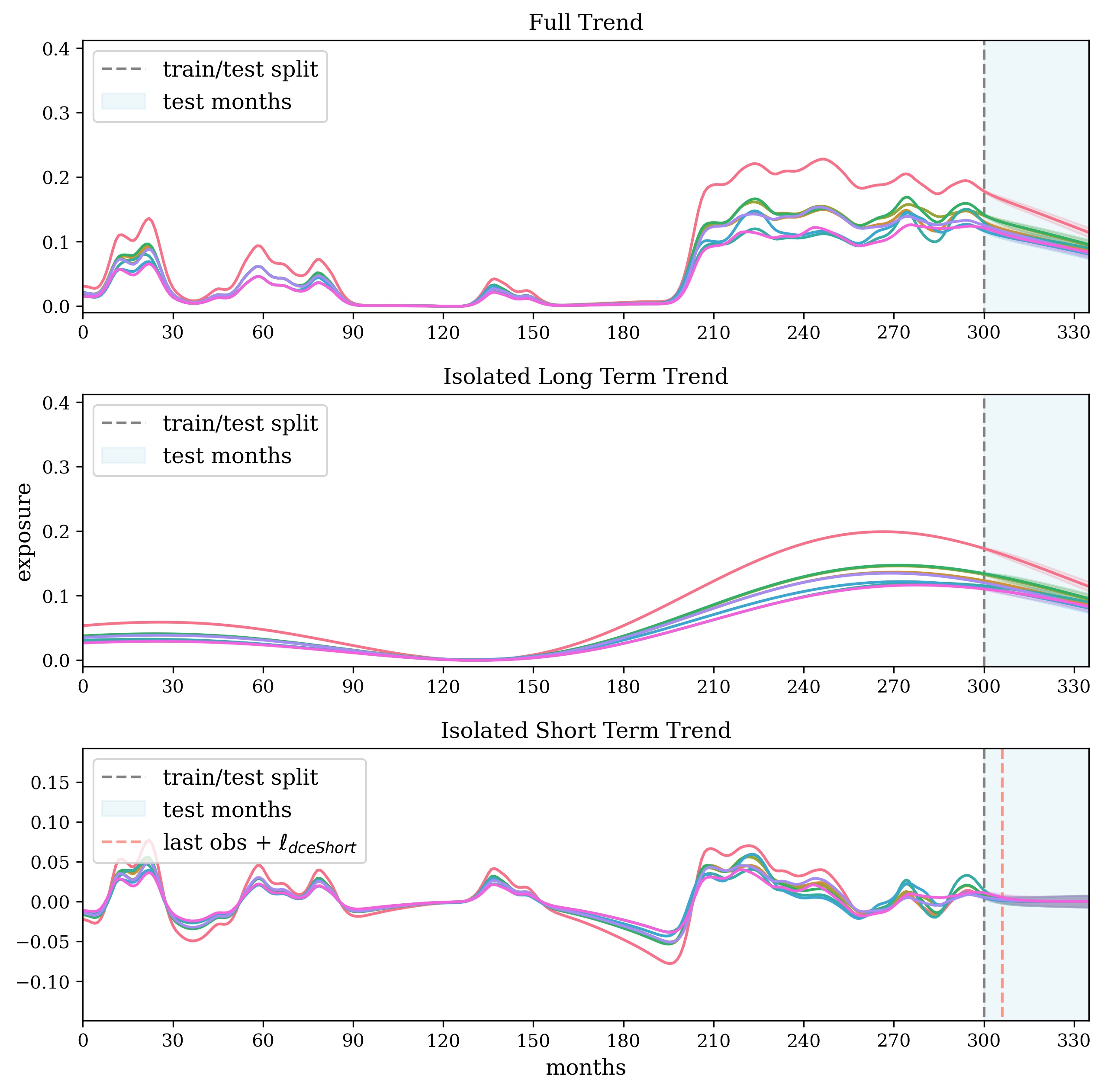}
    \caption{\footnotesize{The same illustrative sample of eight adjacent timelines as presented in \autoref{tce_sampel_eks}. But now the plotted values as $\mu_{TSCE}$, $\mu_{TSCElong}$ and $\mu_{TSCEshort}$. As such, we here see how each timeline is influenced by the other nearby timelines. Once more, the gray dashed line separates the training data from the test data.}}\label{tsce_sampel_eks}
\end{figure}

Using the specified formulation and the estimated hyperparameters, I can now extrapolate $f_{TSCE}$ into the future months $x_{tnew}$ of the test set as already done with $f_{TCE}$. I have plotted eight time-lines in \autoref{tsce_sampel_eks}. First is the full trend $f_{TSCE}$, then $f_{TSCElong}$ and $f_{TSCEshort}$.\par 

Compared to $\ell_{TCEshort}$ we see that $\ell_{TSCEshort}$ reaches relatively longer into the unknown future -- potentially making this feature more relevant for forecasting purposes. Given the relatively short forecasting window the difference between $\ell_{TCElong}$ and $\ell_{TSCElong}$ does not manifest in this plot, but in a operational setting the limit of this length scale means that the spatial information expires after approximately $75$ months, or roughly $6$ years.\par

I can now derive $\mu'_{TSCE}, \mu''_{TSCE}, M_{TSCE}, \mu'_{TSCElong}, \mu''_{TSCElong}, M_{TSCElong}, \mu'_{TSCEshort}, \mu''_{TSCEshort}$, and $M_{TSCEshort}$. The values taken directly from $\mu_{TSCE}*$ are likely relevant since conflict traps are expected to have consequences beyond the cells directly experiencing conflict. The derived values $\mu'_{TSCE}*$ and $\mu''_{TSCE}*$ might prove valuable as they provide signals regarding the tightening or loosening of a broader conflict trap. The cumulative sums $M_{TSCE}*$ might carry relevance since it contains information regarding the combined amount of exposure a cell has experienced throughout all included months.\par

Revisiting the guiding criteria pertaining to the spatial dimension, the present approach fulfils them all. Using a disaggregated geographic grid allows us to analyse and model spatial patterns at arbitrarily high complexity levels, only hampered by the resolution of the data and computational power available. We need not define the number of relevant neighbours since all observations are considered. Furthermore, the deterioration rate of influence is estimated rather than assumed. Lastly, spatial units will be influenced directly by the surrounding patterns and magnitudes of conflict. High magnitude will translate into higher influence, as will encirclement compared to adjacency. Indeed, such properties are part of what compelled \citet{gelfand2016spatial} to declare the combination of spatial data and Gaussian processes a \enquote{beautiful marriage} \citep[86]{gelfand2016spatial}.

%% file: sections/combining.tex
I have now estimated and extrapolated $2$ functions for each cell's timeline: temporal conflict exposure $f_{TCE}$ and tempo-spatial conflict exposure $f_{TSCE}$. Each consists of a long-term trend and a short-term trend: $f_{TCEshort}, f_{TCElong}, f_{TSCEshort}$ and $f_{TSCElong}$. I use the maximum likelihood estimate of functions as features $\mu_{*}$ and derive $\mu'_{*}$, $\mu''_{*}$ and $M_{*}$ as described above. This gives me $24$ features pertaining to the patterns of conflict. An overview is presented in \autoref{overview}.\par

\begin{table}[!htb]
\centering
\caption{\textbf{Features}\\Spatial and temporal conflict patterns as features}\label{overview}
	\begin{tabular}{m{6.7cm} m{3.4cm} m{3.4cm}}
	\toprule
                                            &  \thead{Strictly temporal\\$(TCE)$}        & \thead{Tempo-spatial\\$(TSCE)$}   \\
	\midrule
	Full trend magnitude/exposure   & \thead{$\mu_{TCE}$}                & \thead{$\mu_{TSCE}$} 	        \\
	Full trend slope                & \thead{$\mu'_{TCE}$}               & \thead{$\mu'_{TSCE}$} 	        \\
	Full trend acceleration         & \thead{$\mu''_{TCE}$}              & \thead{$\mu''_{TSCE}$} 	        \\
    Full trend cumulative sum       & \thead{$M_{TCE}$}                  & \thead{$M_{TSCE}$} 	        \\
	\midrule

	Short-term magnitude/exposure   & \thead{$\mu_{TCEshort}$}           & \thead{$\mu_{TSCEshort}$} 	        \\
	Short-term slope                & \thead{$\mu'_{TCEshort}$}          & \thead{$\mu'_{TSCEshort}$} 	        \\
	Short-term acceleration         & \thead{$\mu''_{TCEshort}$}         & \thead{$\mu''_{TSCEshort}$} 	        \\
    Short-term cumulative sum       & \thead{$M_{TCEshort}$}             & \thead{$M_{TSCEshort}$} 	        \\
	\midrule

    Long-term magnitude/exposure    & \thead{$\mu_{TCElong}$}            & \thead{$\mu_{TSCElong}$} 	        \\
	Long-term slope                 & \thead{$\mu'_{TCElong}$}           & \thead{$\mu'_{TSCElong}$} 	        \\
	Long-term acceleration          & \thead{$\mu''_{TCElong}$}          & \thead{$\mu''_{TSCElong}$} 	        \\
    Long-term cumulative sum        & \thead{$M_{TCElong}$}              & \thead{$M_{TSCElong}$} 	        \\
    &&\\

    \bottomrule
	\end{tabular}
\begin{tablenotes}
\item The table contains all $24$ features generated the capture the temporal/spatial patterns of conflict.
\end{tablenotes}
\end{table}

Using the presented approach ensures that the $24$ features span both the training and the test set -- without compromising the out-of-sample design at any point. All patterns a estimated using the training set and then extrapolated into the test set. This is achieved without the use of a sliding-window/one-step-ahead model. The fact that we can use the model specifications together with the estimated hyperparameters and past data to extrapolate patterns of exposure into an unknown future is a key feature of Gaussian processes.\par

As such, any supervised machine learning algorithm can now be directly trained on the training set and evaluated on the test set without applying any leads or lags. To ensure comparability, I use an ensemble of Random Forest models similar to that used in \citet{hegre2019views}. Before training the ensemble, I curate the roster of features created above to find the most relevant subset. I do this for two reasons: I want to survey which kind of exposure is most important and how many features are needed to create satisfying predictions. And I want to pursue a parsimonious framework to minimize overfitting.\par 

\subsection{Feature selection}

To avoid an exhaustive search, I use \emph{forward feature selection} \citep[138-139]{herlau2016introduction}. This algorithm is described in the online appendix.\par  

Here, the forward feature selection algorithm identifies four features as the optimal subset: $\mu_{TCElong}$, $\mu_{TSCEshort}$, $\mu_{TSCElong},$ and $\mu_{TSCE}''$. The first feature selected is $\mu_{TCElong}$. The selection of this feature indicates that the long-term history of local conflict is an important predictor for future conflict. Given our knowledge of conflict traps, this is hardly surprising and fits neatly with theory. The second selection, $\mu_{TSCEshort}$. This selection indicates that short-term spatial variations also contain important information. Given that $\ell_{TSCEshort} = 7.17$ we can deduce that the contribution of this feature lies at the beginning of the forecast but is evidently still consequential enough to warrant selection. The next feature selected is $\mu_{TSCElong}$. This feature is arguably one of the features carrying the broadest signal. It encompasses information on long term tempo-spatial patterns. Again, this selection fits well with insight regarding conflict traps and theory stipulating the importance of spatial conflict diffusion. The last selected feature is $\mu''_{TSCE}$. This selection tells us that the rate of change in nearby conflict levels (full trend) also carries some relevant signal. It is likely that the increase/decrease in spatial conflict patterns provides a signal regarding the regional waxing and waning of conflict traps.\par

\subsection{The Random Forest ensemble}

Having selected four specific features to represent the patterns of conflict, I use a supervised machine learning algorithm that can combine these patterns into final forecasts. To this end, I follow \citet{hegre2019views} and construct an ensemble of $1,000$ Random Forest models. Details can be found in the online appendix.\par

%% file: sections/evaluation.tex
To evaluate the out-of-sample predictive performance of my approach, I utilize the test set along a roster of suitable metrics introduced below. When applicable the ViEWS early-warning-system \citep{hegre2019views} will be used as comparison.\par

\subsection{Principle metrics}
Following \citet{hegre2019views}, the primary evaluation metric used is the \emph{precision-recall} ($PR$) curve and the related metrics \emph{recall}, \emph{precision} and the \emph{average precision} ($AP$) score\footnote{\citet{hegre2019views} reports \emph{Area under the $PR$-curve} ($AUPR$). $AP$ is an unbiased estimation of this metric and they are thus directly comparable \citep{boyd2013area, su2015relationship}.}. I also use the \emph{Receiver Operating Characteristic} ($ROC$) curve and the \emph{Area Under the Curve} ($AUC$) score \citep{Friedman_2001, He_2008}. Using multiple metrics is warranted, since model performance is inherently multi-dimensional \citep[165]{hegre2019views}. For more substantially evaluations and visualisations I also use True Positives $TP$, False Positives $FP$, True Negative $TN$ and False Negative $FN$. A more comprehensive introduction to metrics can be found in the online appendix.\par

\subsection{Comparative performance}

\begin{figure}[!htb]
	\centering
	\includegraphics[scale=0.6]{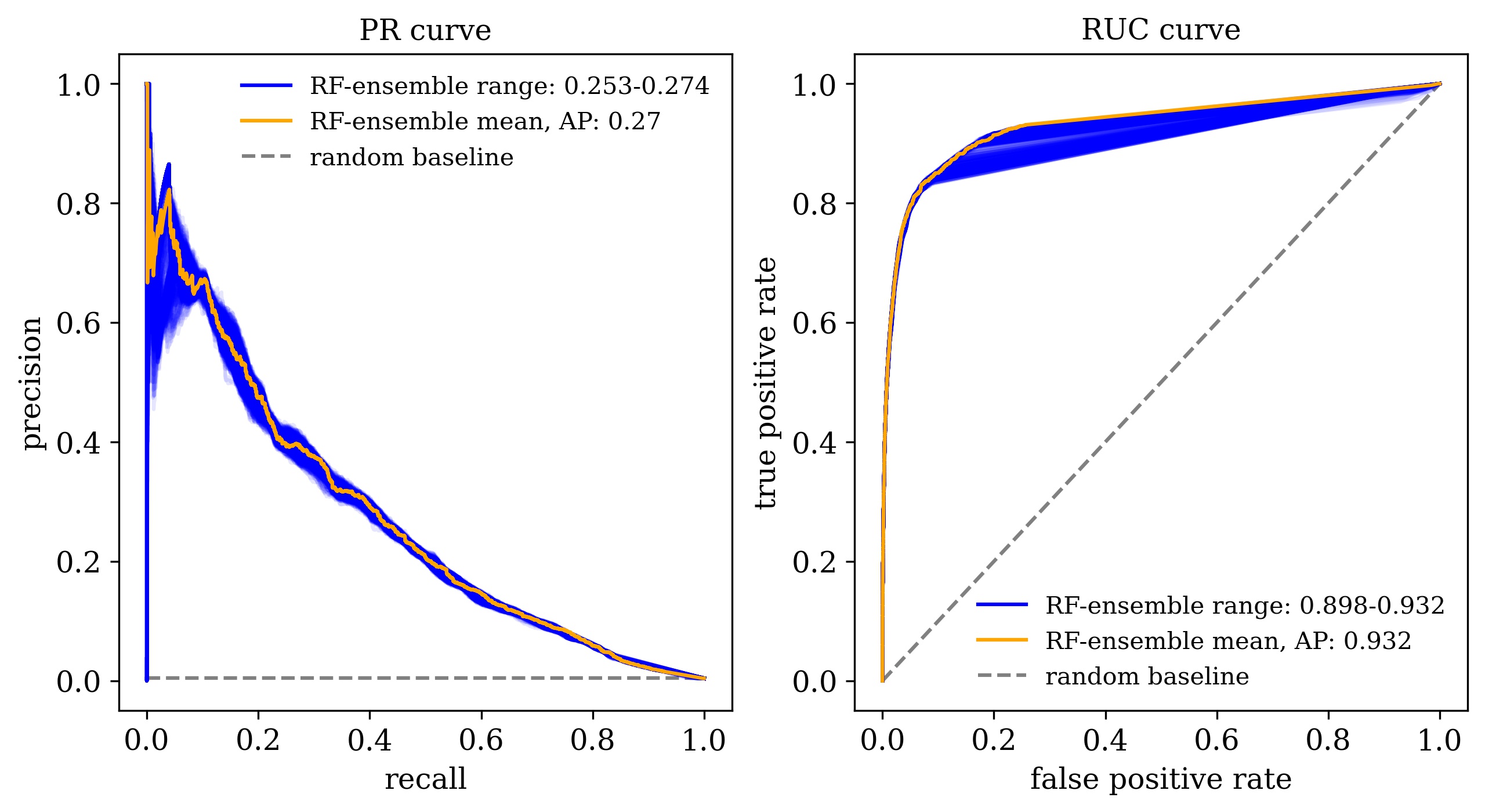}
    \caption{\footnotesize{PR curve/AP score and ROC Curve/AUC score based on $1,000$ models. Note that the \enquote*{ensemble} is not the mean of the individual AP/AUC scores, but the result of using the mean of the individual predictions as a point estimate for the probability of conflict. The gray dashed lines represent models doing no better than random.}}\label{curves}
\end{figure}

The performance of my approach on the full test set is presented in \autoref{curves}. Blue lines depicts the performance of the individual models while the orange line shows the performance of the ensemble, which achieves an $AP = 0.2704$ and an $AUC = 0.9318$.\par

In \autoref{views_comp} I show my results compared to those presented in \citet{hegre2019views}. Note that my ensemble still only uses the four features on conflict exposure selected above. No other features are included and as such the ViEWS component most readily comparable with my approach is the \enquote*{conflict history component}. Surveying the results we see that my approach is able to outperform all models save their best ensemble -- and even this full early-warning-system only outperforms my approach by a very slight margin: $0.0166$ $AUC$ and $0.007$ $AP$.\par

\begin{table}[!htb]
\centering
\caption{\textbf{Comparison with ViEWS}\\Out-of-sample predictions, 36 months January 2015 –- December 2017}\label{views_comp}
	\begin{tabular}{m{7.1cm} m{3.2cm} m{3.2cm}}
	\toprule
                                                                                            &  \thead{AUC}      & \thead{AP}    \\
	\midrule
	VIEWS Baseline                                                                          & \thead{0.6324}    & \thead{0.049}   \\
	VIEWS Baseline + conflict history theme                                                 & \thead{0.8920}    & \thead{0.225}   \\
	VIEWS Baseline + conflict history theme + social and natural geographic themes          & \thead{0.9225}    & \thead{0.227}   \\
    VIEWS All themes                                                                        & \thead{0.9125}    & \thead{0.245} 	\\
    \rowcolor{gray} VIEWS Ensemble                                                          & \thead{0.9484}    & \thead{0.277} 	\\
    \rowcolor{lightgray} Presented approach (temporal/spatial signal)                       & \thead{0.9318}    & \thead{0.270} 	\\
    &&\\
  
    \bottomrule
	\end{tabular}
\begin{tablenotes}
\item The presented approach uses four features, $\mu_{TCElong}$, $\mu_{TSCEshort}$, $\mu_{TSCElong}$ and $\mu_{TSCE}'$, together in a simple ensemble of Random Forest models. All results pertaining to ViEWS are from \citet{hegre2019views}.
\end{tablenotes}
\end{table}

\begin{figure}[!htb]
	\centering
	\includegraphics[scale=0.6]{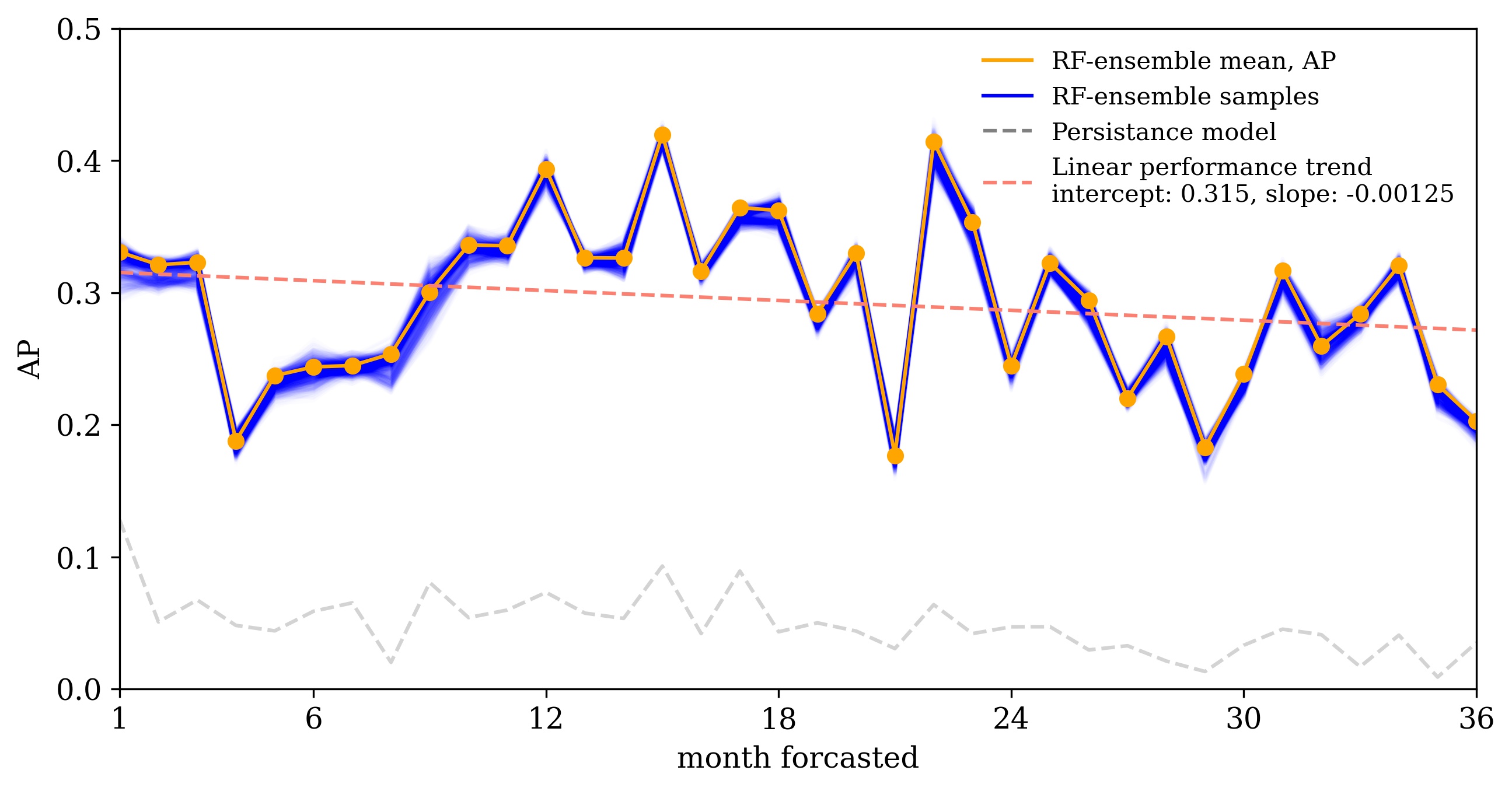}
    \caption{\footnotesize{AP results for the ensemble through the forecast months. The \enquote*{Persistence model} is a simple baseline created by using the last observation in a given cell as the prediction for all future conflicts in said cell. The corresponding results from ViEWS can be found in \citet{hegre2019views} p. 167, as I have not been able to obtain this data.}}\label{AP_trends}
\end{figure}

Looking at the performance across months in \autoref{AP_trends}, we see that my approach outperforms ViEWS' best ensemble during the first three months of the forecast. This highlights just how much predictive power past patterns contain, especially in the short run.\par

As already noted, I am not suggesting my approach as an alternative to full-scale early-warning systems such as ViEWS. The comparison is simply meant to showcase how much potential predictive power we might extract from past conflict patterns. My contribution should be understood as one component focused on tempo-spatial conflict exposure, and as such similar to ViEWS \enquote*{conflict history theme}. Such a component might be incorporated into a larger model including features regarding social, economic, demographic, political, and geographical themes -- themes which likely contain information distinct from what I here capture, and thus a union might generate even more encouraging results. The predictive power of the approach also indicates that the features are good representations of conflict exposure, which makes them prudent controls given efforts focusing on estimation and causal identification.\par 

\subsection{Substantial evaluation}

To get a more substantial assessment of the approach's performance I now present a \emph{confusion map} showing the distribution of $TPs$, $FPs$, $TNs$ and $FNs$ across a number of curated test months\footnote{A time-lapse over all months can be found at \url{https://github.com/Polichinel/the_currents_of_conflict/tree/main/time_lapse}}.\par

\begin{figure}[!htb]
	\centering
	\includegraphics[scale=0.6]{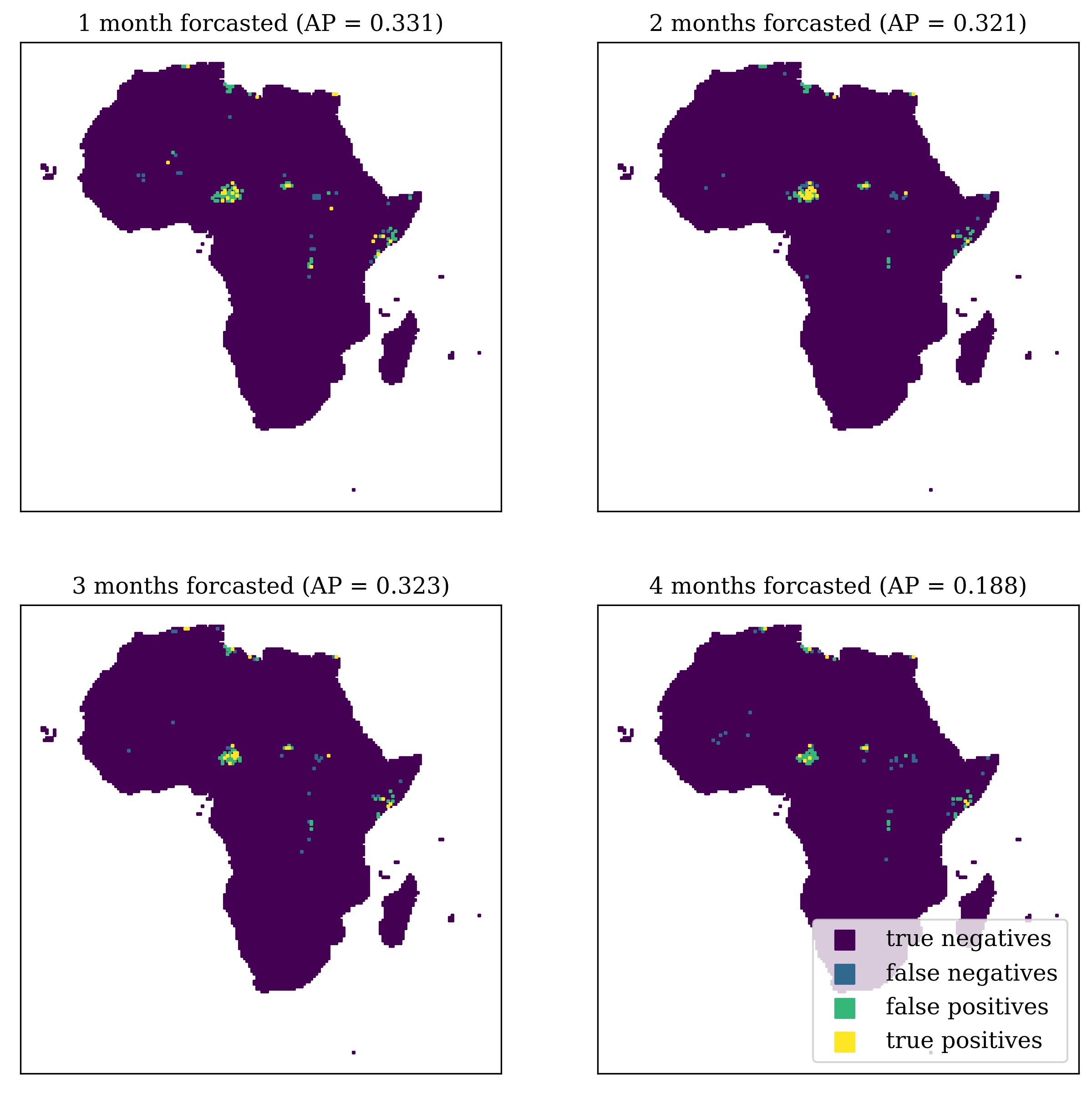}
    \caption{\footnotesize{Illustrative example of confusion maps from the first four months of the test data (January, February, March and April $2015$). The threshold is set at $0.06$ to obtain roughly the same amount of conflicts as observed in the last training month.}}\label{confusion_map}
\end{figure}

To obtain $TPs$, $FPs$, $TNs$ and $FNs$ I need to convert the predicted probabilities to binary classification $y_{binary} \in \{0,1\}$. This requires some threshold dictating when a prediction will be classified as an event versus a non-event. I set a threshold that ensures that I predict roughly the same number of conflicts as was observed in the last month of the training set. Given this threshold, confusion maps pertaining to the first four months in the test set forecast are presented in \autoref{confusion_map}.\par

I visualize these four months because we see here three months where my approach achieves rather good results, and then a fourth month with drastically worse results. As such this is a good opportunity to asses strength and weaknesses of my approach. Looking at \autoref{confusion_map}, the drop in performance can be traced to two developments. Firstly, the spread of conflict in Mali and along the border between Sudan and South Sudan increases the number of $FN$. Secondly, a decrease in conflict in northeastern Nigeria leads to an increase in $FP$. The development in northeastern Nigeria is a product of a hard offensive against Boko Haram initiated in February $2015$ (the $2^{nd}$ month of the forecast) by Nigerian forces. By April, Boko Haram was under heavy pressure and mainly confined to the Sambisa Forest \citep{CoronesAfrican}. We already know that fast developments are a major challenge in conflict forecasting \citep{hegre2021can}. As illustrated, this is one challenge my approach does not manage to address.\par

It should be noted that $FP$s might be less problematic than $FN$s in an operational setting. That is, even if conflict does not manifest in a given cell, the risk of conflict in said cell might still be genuine. This is particularly true when we have highly disaggregated tempo-spatial units such as PRIO-grid-months. Indeed, the increased predictive power seen in later test-set months is partly due to renewed violence in northeastern Nigeria. This indicates that the retreat of Boko Haram in $2015$ should be understood as part of a short-term trend -- the long-term risk of conflict never waned. The danger of $FP$s is that resources might be allocated to areas where no conflict is likely to happen. However, if the $FP$s tend to indicate high-risk zones where conflict merely failed to manifest, resources allocated here might not constitute waste after all. Not predicting conflicts where conflicts do erupt, i.e., the generation of $FN$s, is a more pressing issue.\par
    
\subsection{Estimating probability or magnitude}

While I here exemplify the potential of my approach on a classification task, the approach is also well suited for regression tasks. That is forecasting the magnitude of conflicts instead of, or along with, the probability of conflict. Indeed, I would argue that conflict forecasting is inherently a regression task and not a classification task. When using data from sources such as the UCDP, we do have a countable target: Conflict fatalities ($y \in \mathbb{W}$, potentially logged). The \enquote*{binaryfication} creating a feature denoting conflict/non-conflict is artificial. We apply this transformation because we want to estimate the probability of conflict, but forecasting the magnitude of future conflicts appears just as paramount. Perhaps even more so when generating highly disaggregated forecasts, as policy makers and actors now have the opportunity to formulate very specific strategies. 

The challenge, however, is that no truly appropriate off-the-shelf metrics or benchmarks exist yet that might be used to evaluate such efforts. Do to the rarity of conflict events and the tempo-spatial relations between such events, conventional metrics such as \emph{mean squared error} or \emph{absolute squared error} appear imprudent. As such this is an avenue I suggest future efforts relegate some attention.\par

%% file: sections/conclusion.tex

To facilitate valid evaluation, I based this effort on the replication data from the current state-of-the-art framework ViEWS \citep{hegre2019views}. I use the same data, the same temporal and geographical units (monthly PRIO-grid cells), the same subset (state-based conflicts in Africa), the same training/validation/testing splits ($1990-2011$, $2012-2014$, and $2015-2017$), and a similar machine Learning ensemble for the final forecast 

The contribution of this article is a novel approach capable of estimating and forecasting future conflict patterns using only past conflict patterns. I show how we might use temporally and spatially disaggregated data on conflict-events in tandem with Gaussian processes to estimate and extrapolate features pertaining to the spatial and temporal patterns of conflict. This approach also enables me to identify and extract long- and short-term trends. From these trends, I extract even more information regarding the patterns of conflict by deriving the slope, acceleration, and cumulative mass of the trends. All in all, I generate $24$ features pertaining to the patterns of tempo-spatial conflict exposure. Of these, four features are chosen via forward feature selection to be part of the final out-of-sample evaluation.\par 

The final four features, denoted $\mu_{TCElong}$, $\mu_{TSCEshort}$, $\mu_{TSCElong}$ and $\mu_{TSCE}''$, encompasses a mixture of spatial and temporal signals. The use of Gaussian processes allows me to extrapolate these features into the \enquote*{future months} of the test set without compromising the out-of-sample design and without leading or lagging any variables. Thus, any supervised machine learning model(s) can be trained on the training set and evaluated on the test set without using sliding window/one-step-ahead models.\par

Another attractive property of the approach presented here is the fact that we get valid estimates of data uncertainty. As such we can easily asses how the uncertainty around different kinds of conflict exposure proliferates as we move forward in time through our forecasts. And indeed, the estimated length scale hyper-parameters can be understood as heuristic guide to the temporal limit of our forecasting ability. Thus, revealing both theoretically and practically relevant insights.\par

The approach also exhibits great predictive power: With an $AP = 0.2704$ and an $AUC = 0.9318$ on a $36$ month forecasting window, I find that my approach outperforms ViEWS' comparable and thematically similar conflict history component. Indeed, my approach outperforms all but ViEWS' best and final ensemble on a $36$ month forecast average. That the best and final ViEWS model outperforms my approach is hardly surprising -- it is, after all, a far more comprehensive, fully operational, early warning system. What is somewhat surprising is that it only outperforms my approach by $0.0166$ on the $AUC$ score and $0.007$ on the $AP$ score. Furthermore, I find that the approach presented in this article does manage to outperform ViEWS' final ensemble during the first three months of the forecast. Notably, this is achieved with four features pertaining to conflict temporospatial exposure, and a Random Forest ensemble.\par

Nevertheless, my contribution should still only be seen as a way to create one specific component for a more comprehensive early-warning system -- such as ViEWS. A component capturing the temporal and spatial patterns of conflict exposure. While the component could be used alone as a \enquote*{cheap} alternative to larger systems, incorporating it into a framework which also includes information on structures, demography, and institutions would likely increase prediction power markedly.\par

Apart from parsimony and state-of-the-art performance, my approach exhibits a number of other benefits: As it does not require any sliding window or one-step-ahead modeling, I do not lose any data by having to \enquote*{lead} or \enquote*{lag} features. I avoid ad hoc function specifications since trends and diffusion patterns are estimated directly from the data. The fact that I am estimating patterns also means that this approach generates substantial insights regarding temporal and spatial patterns of conflict. Furthermore, the estimated patterns are suitable not just for estimating conflict probabilities but also for estimating future conflict magnitudes, i.e., regression tasks. Lastly, the generated features can easily be used in efforts focusing on parameter estimation or causal identification as control variables for conflict traps and spatial diffusion.\par

%% file: sections/replication.tex
For replication, the ViEWS data can be found at \hyperlink{http://views.pcr.uu.se/downloads/.}{http://views.pcr.uu.se/downloads/.} and the PRIO grid shape file can be found at \hyperlink{https://www.prio.org/Data/PRIO-GRID/}{https://www.prio.org/Data/PRIO-GRID/}. All code is written with Python 3 and supplied for review in a .zip file. The code can also be accessed at the author's GitHub page \url{https://github.com/Polichinel/the_currents_of_conflict}.\par

%% file: sections/acknowledgements.tex
This paper draws on research originally conducted as part of the author’s Ph.D. dissertation at the University of Copenhagen. The author is grateful for the guidance and support of supervisors Lene Hansen and Jacob Gerner Hariri, and thanks the dissertation committee — Frederik Hjorth, Lisa Hultman, and Nils Weidmann — for valuable feedback.

Subsequent development and revision of this work have been supported by ongoing research at the Peace Research Institute Oslo (PRIO). The author is particularly thankful for discussions with colleagues and collaborators at PRIO and beyond, some of whom may be added as co-authors in future versions.

This research has received support from multiple projects: the dissertation was developed with funding from Bodies as Battleground: Gender Images and International Security (Independent Research Fund Denmark, PI: Lene Hansen), while the current revision is supported by Societies at Risk: The Impact of Armed Conflict on Human Development (Riksbankens Jubileumsfond, PI: Håvard Hegre).